\documentclass{article}
\usepackage{spconf,amsmath,graphicx}
\usepackage{array}
\usepackage{multirow}
\usepackage{subfigure}
\usepackage{color}

\title{Exploitation of Semantic Keywords for Malicious Event Classification}

\name{
\textsuperscript{$\star$}Hyungtae Lee$^{\dagger\ddagger}$,
\textsuperscript{$\star$}Sungmin Eum$^{\dagger\ddagger}$,
\textsuperscript{$\star$}Joel Levis$^{\mathsection}$,
Heesung Kwon$^{\ddagger}$,
James Michaelis$^{\ddagger}$,
and Michael Kolodny$^{\ddagger}$
\thanks{\textsuperscript{$\star$}These authors contributed equally to this work}
}

\address{$^{\dagger}$Booz Allen Hamilton Inc., McLean, Virginia, U.S.A.\\
$^{\ddagger}$U.S. Army Research Laboratory, Adelphi, Maryland, U.S.A.\\ $^{\mathsection}$Ohio University, Athens, Ohio, U.S.A.\\ }

\begin{document}
\maketitle
\begin{abstract}

Learning an event classifier is challenging when the scenes are semantically different but visually similar. However, as humans, we typically handle such tasks painlessly by adding our background semantic knowledge. Motivated by this observation, we aim to provide an empirical study about how additional information such as semantic keywords can boost up the discrimination of such events. To demonstrate the validity of this study, we first construct a novel Malicious Crowd Dataset containing crowd images with two events, benign and malicious, which look visually similar. Note that the primary focus of this paper is not to provide the
state-of-the-art performance on this dataset but to show the beneficial aspects of using semantically-driven keyword information. By leveraging crowd-sourcing platforms, such as Amazon Mechanical Turk, we collect semantic keywords associated with images and then subsequently identify a subset of keywords (e.g. police, fire, etc.) unique to specific events. We first show that by using recently introduced attention models, a na\"ive CNN-based event classifier actually learns to primarily focus on local attributes associated with the discriminant semantic keywords identified by the Turks. We further show that incorporating the keyword-driven information into early- and late-fusion approaches can significantly enhance malicious event classification.





\end{abstract}

\section{Introduction}




Images associated with very different events can often be represented with similar visual attributes, as shown in Figure 1. In Figure 1, both images include crowds as main foreground attributes, yet we can immediately discern that the two images contain drastically different semantic events. The content of the right image is ``malicious'' as opposed to the left image being ``benign'' due mainly to some specific objects contained in the image, such as fire, smoke, police, etc. The above observation indicates that exploiting relevant local attributes (local objects) jointly with global attributes (the whole scene) is quite essential to better discriminating the events containing complex events. 
What is more important is being able to find out which objects are more correlated (semantically meaningful) than others with respect to a certain set of events.

\begin{figure}[t]

\begin{minipage}[b]{1.0\linewidth}
  \centering
  \centerline{\includegraphics[width=\linewidth,trim=5mm 20mm 5mm 15mm]{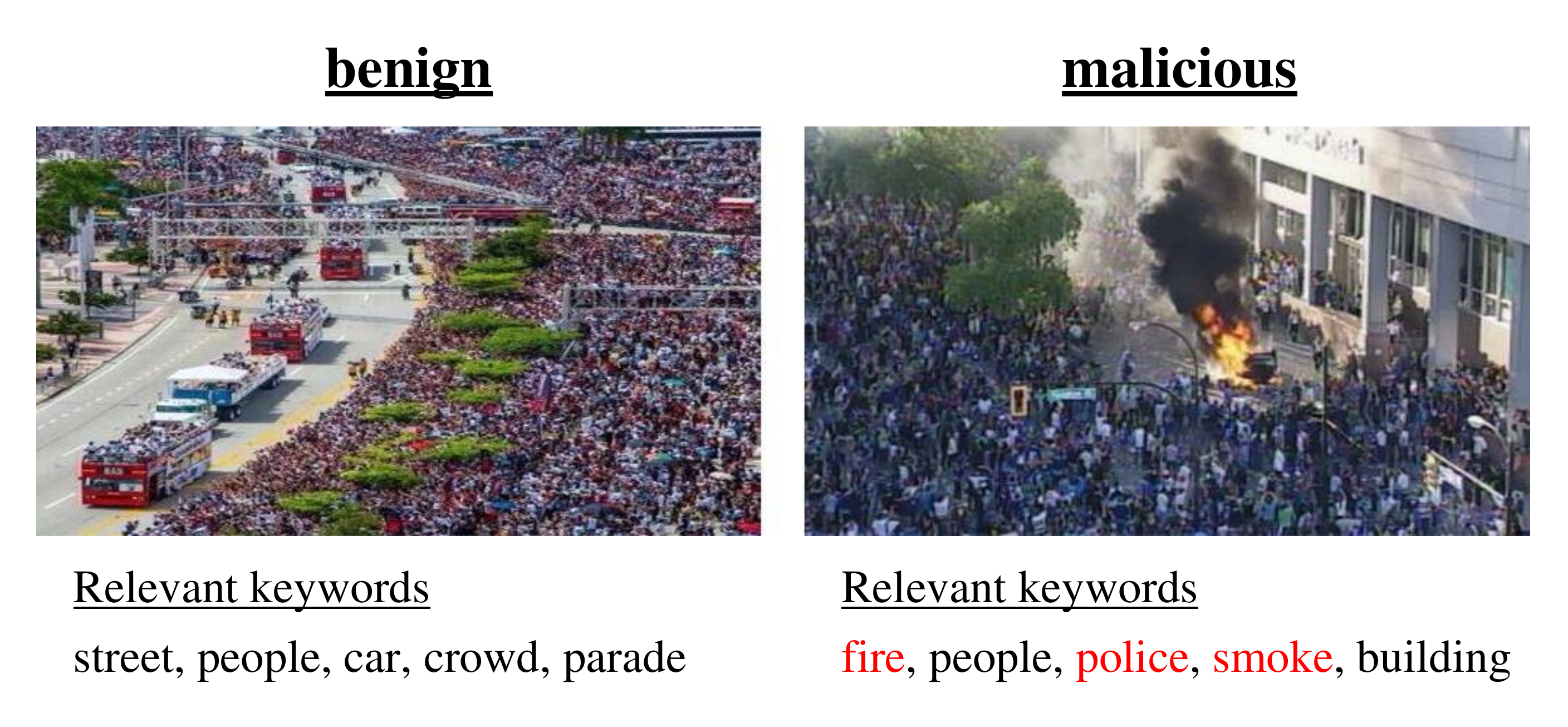}}
\end{minipage}

\caption{A pair of similarly looking crowd images from different events with relevant object contents.}
\label{fig:intro}
\end{figure}

In this paper, we introduce a new event dataset and then conduct a study, which verifies our argument that identifying highly relevant objects to associated events are crucial to improving classification performance. We first collect a set of images which contain two visually similar, yet semantically different events: benign and malicious. Since most benchmark datasets \cite{LLiICCV07,SOhCVPR11,2016trecvidawad} collected for event classification do not deal with this problem, our newly constructed dataset will be useful for the community. We also collect a number of keywords that appear in each  image in the dataset, as listed below each image in Figure \ref{fig:intro}. We asked the users on the Amazon Mechanical Turk to describe the semantic contents of each image in terms of keywords without providing the semantic event labels. Then we select non-overlapping distinctive and frequently occurring keywords for the malicious event, which we aim to identify and treat them as the representative ``semantic keywords''.

\begin{figure*}[t]
\includegraphics[width=\linewidth,trim=5mm 30mm 5mm 10mm]{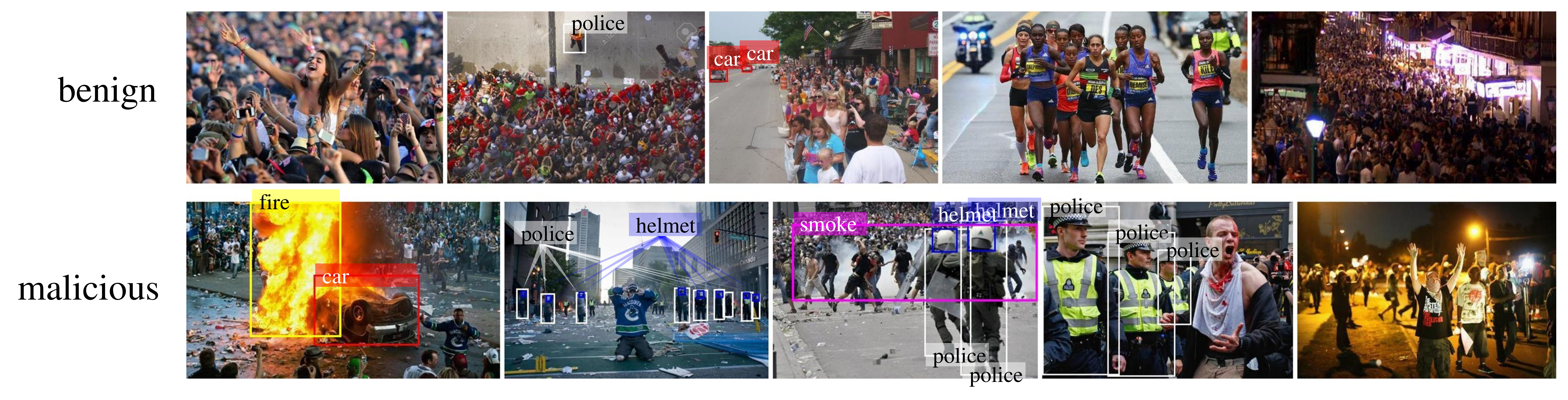}
\caption{{\bf Malicious Crowd Dataset.} Example images for the benign and malicious events are shown. The groundtruth bounding box information which corresponds to the selected malicious semantic keywords is also included in the dataset. Yellow, purple, blue, white, and red boxes correspond to fire, smoke, helmet, police, and car, respectively.}
\label{fig:images}
\end{figure*}

Secondly, we analyze how these human-driven sets of semantic keywords align with the visual attributes inherently learned by the na\"ive event classifier. For this analysis, we leveraged the visualization approach based on the top-down neural attention maps \cite{JZhangECCV16} which indicate the regions on which the deep CNN-based event classifier mainly focuses. These ``machine-driven'' attention maps show that they clearly share high relevance with the ``human-driven'' semantic keywords although the classifier is not supported with any additional semantic information in the learning process.

Lastly, we carry out a study to verify the practicality of explicitly incorporating these semantic keywords using various fusion approaches, which include a novel CNN-based architecture (IOD-CNN) developed by part of the authors \cite{SEumICIP17}. We show that these keyword-driven information is effective in helping out the event classification task regardless of whether the information is used in an early- or late-fusion scheme.


Our contributions are summarized as follows:
\begin{enumerate}

\item We introduce a new Malicious Crowd Dataset containing semantically different malicious/benign images, crowd-sourced semantic keywords, and groundtruth bounding box annotations for the objects corresponding to the semantic keywords.

\item By analyzing the attention maps, we provide valuable indications that even na\"ive malicious event classifier learns to focus on the regions which align with the semantic keyword set.

\item We provide the community with novel findings based on the empirical study, which verifies the practicality of explicitly using the semantic keyword information for malicious event classification.

\end{enumerate}


\section{Malicious Crowd Dataset}
\label{sec:dataset}

We have constructed the Malicious Crowd Dataset which consists of images along with the malicious/benign event labels. It also includes the malicious semantic keywords and their corresponding bounding box information for each image.

\subsection{Malicious and Benign Crowd Images}
\label{ssec:images}

The dataset contains 1133 crowd images equally split into two classes: benign and malicious.  A benign image would portray a ``non-alarming'' scene, while a malicious one would be alarming and potentially dangerous.

The images were collected from the web using various search terms.  For benign images, search terms such as marathon, pedestrian, crowd, parade, and concert were used.  Terms such as riot and protest were used to gather the malicious crowd images.  
Figure~\ref{fig:images} illustrates some example images from each class.

\subsection{Semantic Keywords}
\label{ssec:semanticKeywords}

\noindent{\bf Malicious Semantic Keywords Selection.}
To describe the contents of each of the crowd images, Amazon Mechanical Turk was used.  A human was responsible for assigning five keywords to each image based on what objects are observed within.  To ensure the accuracy of the Mechanical Turk results, we manually removed the keywords which were incorrectly assigned.

\begin{figure}[t]

\begin{minipage}[b]{1.0\linewidth}
  \centering
  \centerline{\includegraphics[width=\linewidth,trim=5mm 25mm 5mm 10mm]{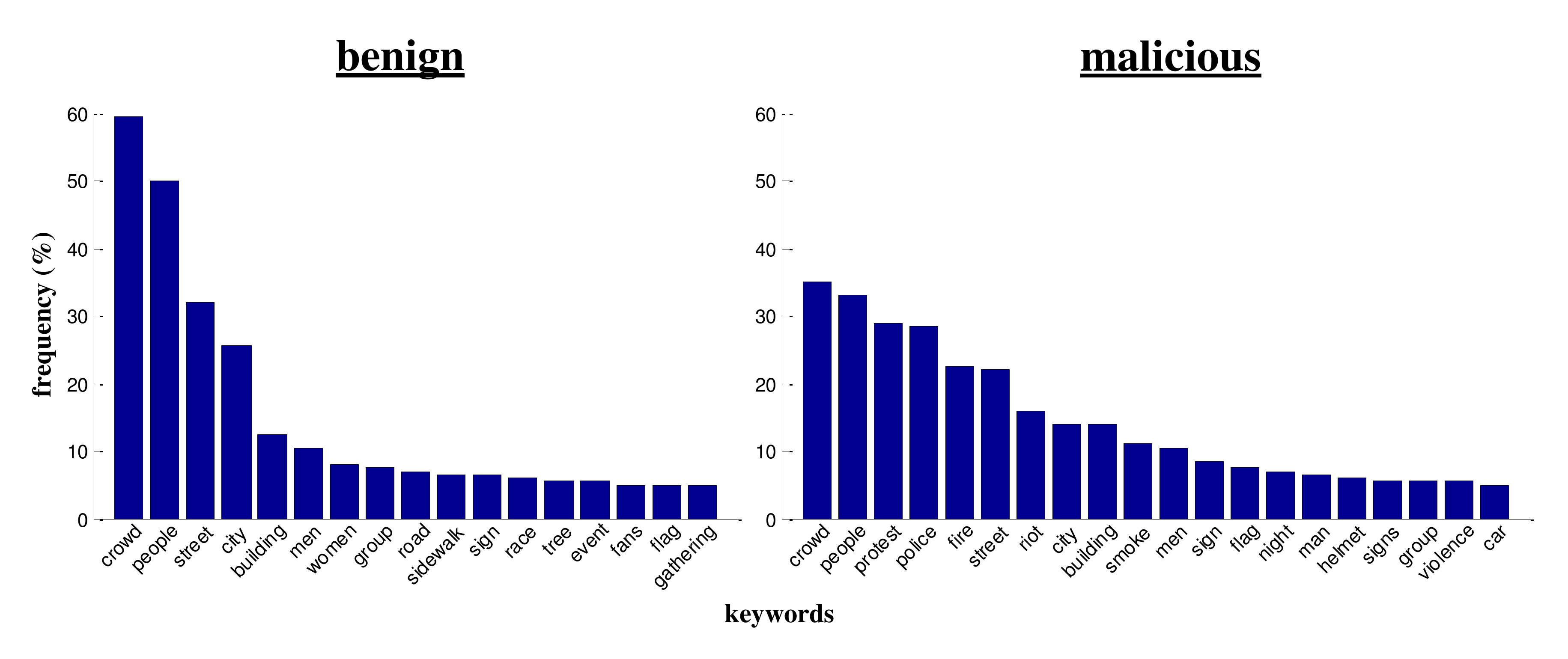}}
  \label{fig:hist_keywords}
\end{minipage}

\caption{Histograms of Relevant Keywords}
\label{fig:frequency}
\vspace{-1.0em}
\end{figure}

\begin{figure*}[t]

\begin{minipage}[b]{1.0\linewidth}
  \centering
  \centerline{\includegraphics[width=\linewidth,trim=10mm 20mm 10mm 5mm]{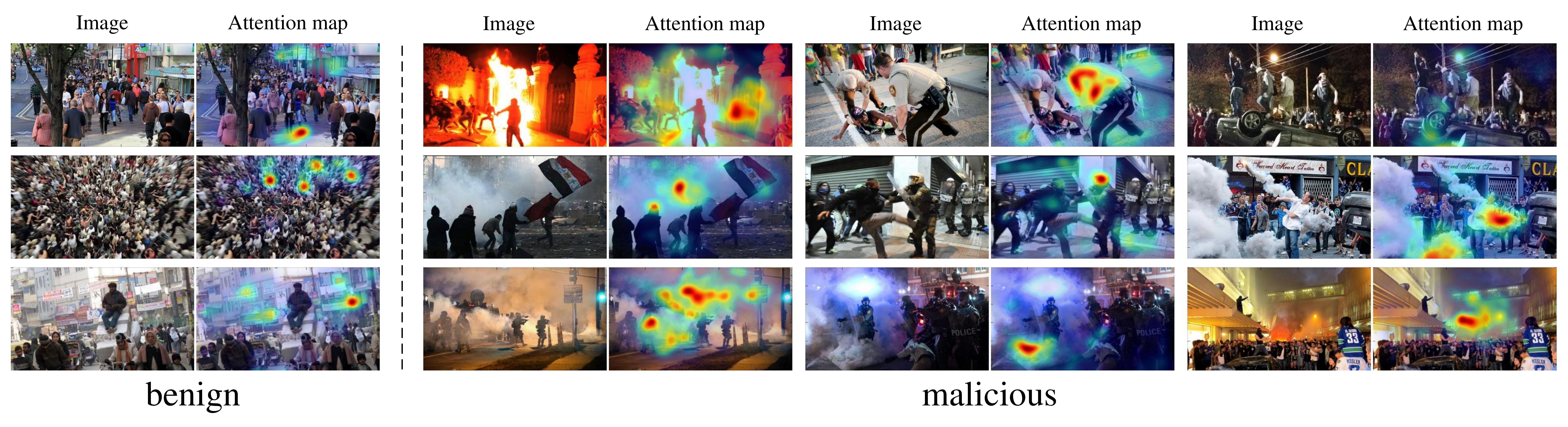}}
\end{minipage}

\caption{{\bf Example Images of Attention Maps.} Na\"ive event classifier is used. Red indicates strongly fired-up regions.}
\label{fig:attentionMaps}
\end{figure*}


After successfully collecting the crowd images and corresponding keywords, identifying keywords only relevant to the malicious class was necessary.  We then constructed two keyword sets, each acquired by selecting the most frequently appearing keywords in the two given classes.  In practice, words that are commonly annotated in 5\% or more images in each class were selected.  As a result of this thresholding, the numbers of selected words for the benign and malicious classes are 17 and 20, respectively.  Selected words and those frequency for both classes can be seen in Figure~\ref{fig:frequency}.  We have refined the set by eliminating the common keywords that appear in both classes.  This elimination resulted in nine malicious keywords. We further eliminated keywords indicating particular phenomena such as {\it protest}, {\it riot}, {\it night}, and {\it violence}. Then {\it police}, {\it fire}, {\it smoke}, {\it helmet}, and {\it car} are included in the final set of malicious semantic keywords. \\

\noindent{\bf Annotation.} After finalizing the set of malicious semantic keywords, to provide the ground truths, we went through all the images and manually labeled the bounding boxes for the five objects which correspond to those keywords.

Table~\ref{tab:num_of_img} shows the number of images where each keyword (object) actually appears.  While {\it police}, {\it fire}, {\it smoke}, and {\it helmet} seem to be closely associated with the malicious event, {\it car} is seen in both events with a similar frequency.  Note that the numbers in the table do not necessarily match the histogram of malicious semantic keywords obtained from crowd-sourcing.  For example, {\it police} appears in 205 out of all 576 malicious images at a rate of 35.59\%, but is assigned only to 28.50\% of the malicious image by the Turks.  This is because the visual contents associated with these keywords are not overly notable in several images.  We can observe that the frequencies of the selected semantic keywords show a notable gap between the two classes, indicating that the purpose of the proposed keyword selection process is achieved.

\begin{table}[t]
\setlength{\tabcolsep}{5.2pt}
\begin{center}
\caption{Number of images where each keyword relevant to the malicious image appears.}
\begin{tabular}{l|c|ccccc}
\hline
class & images & police & fire & smoke & helmet & car \\\hline\hline
benign & 557 & 8 & 1 & 2 & 7 & 57 \\
malicious & 576 & 205 & 144 & 150 & 206 & 65 \\\hline
\end{tabular}
\label{tab:num_of_img}
\end{center}
\end{table}

\section{Visualizing the attention maps for malicious events}
\label{sec:visualization}

As mentioned in Section \ref{sec:dataset}, the final set of semantic keywords is selected by carrying out a process with the human in the loop which may bring certain heuristics. In order to analyze how much these human-driven additional semantics are correlated with the actual event classifier, we have visualized the attention maps which depict the ``excited'' or ``fired-up'' regions on the images learned by the event classifier. To visualize the attention maps, we have used the top-down neural attention model \cite{JZhangECCV16} which uses the excitation back-propagation scheme. We have trained a deep CNN-based event classifier (`Event CNN' in Table \ref{tab:overallComparison}) to be used for generating the attention maps. Example attention maps along with the original images are shown in Figure \ref{fig:attentionMaps}.

We can observe that this na\"ive event classifier (`Event CNN'), even without the additional semantic information, has learned to focus on most of the selected keyword-driven objects. This verifies that the selected semantic keywords and their corresponding objects are inherently being used in distinguishing the malicious events from the benign ones.

\section{Effectiveness of semantic keywords}

\subsection{Evaluation Protocol}
The Malicious Crowd Dataset consists of 1133 images - 576 of 1133 are labeled as the malicious image and the rest are labeled as benign. The dataset is randomly divided into train and test sets which include 905 and 228 images, respectively. Average precision (AP) is used as an evaluation metric.

\subsection{Methodology}
We demonstrate the effectiveness of semantic keyword-driven information by using i) late fusion of classifiers or ii) IOD-CNN (Integrated Object Detection CNN) \cite{SEumICIP17}, which can be considered an early fusion.
\\

\begin{table*}[t]
\setlength{\tabcolsep}{5.6pt}
\caption{{\bf Malicious Event Classification Accuracy.} Late fusion approaches}
\begin{center}
\begin{tabular}{l|c|ccccc|ccccccc}
\hline
 & Baseline & \multicolumn{5}{c|}{Keyword-driven object} & \multicolumn{6}{c}{Late fusion} \\\cline{3-14}
 & (EventCNN) & police & fire & smoke & helmet &  car  & SVM-rbf & DBF & SVM-lin & kNN & LD & LR & EC \\\hline\hline
AP &  72.2 & 58.6 & 56.3 & 68.9 & 53.2 & 49.1  & 74.2 & 75.7 & 75.8 & 75.8 & 76.0 & 76.3 & \underline{{\bf 77.1}} \\\hline
Gain & $\cdot$ & $\cdot$ & $\cdot$ & $\cdot$ & $\cdot$ & $\cdot$ & +2.0 & +3.5 & +3.6 & +3.6 & +3.8 & +4.1 & +4.9 \\\hline
\end{tabular}
\end{center}
\label{tab:ap}
\vspace{-1.0em}
\end{table*}

\noindent{\bf Late Fusion of Event/object Classifiers.} 
We have trained five different object classifiers or detectors which correspond to five selected keywords. For the rigid objects ({\it helmet}, {\it police}, and {\it car}), we have used the deformable part model (DPM), while two separate CNN classifiers were trained for non-rigid objects ({\it fire} and {\it smoke}). A CNN classifier (`Event CNN') for event classification was also trained. All the three CNN classifiers were fine-tuned from \cite{AKrizhevskyNIPS12}.

A late fusion was performed on the output of these six streams. This is to enhance the performance of the baseline event classifier. We tested several fusion methods which include Linear Discriminant Analysis (LD) \cite{RFisherAE10}, Logistic Regression (LR) \cite{DFreedman09}, Support Vector Machines (SVM) \cite{CCortesML95}, $k$-Nearest Neighbor Classifiers ($k$NN) \cite{NAltmanAS92}, Subspace-based Ensemble Classifiers (EC) \cite{THoPAMI98}, and a Dynamic Belief Fusion (DBF) \cite{HLeeWACV16}. For SVM, we used two different kernels which are a linear kernel (SVM-lin) and RBF kernel (SVM-rbf).  
\\

\noindent{\bf IOD-CNN.}
Eum et al. \cite{SEumICIP17} introduced a unified deep CNN architecture which integrates architecturally
different, yet semantically-related object detection
networks to boost event recognition
task. This architecture allows the within-network sharing of the convolutional/fully connected layers across event recognition and object detection tasks. This approach can be considered as an ``early fusion'' which can be differentiated with the ``late fusion''. As the network is learned in an end-to-end fashion, the training can be performed efficiently.

\subsection{Experiments}

Note that the purpose of this paper is not to provide the state-of-the-art performance on this dataset but to provide an empirical study which shows the beneficial aspects of using additional information which are semantically-driven.
\\

\noindent{\bf Late Fusion of Event/object Classifiers.}
Table~\ref{tab:ap} shows the performance for the baseline model (`Event CNN'), keyword-driven object detectors/classifiers, and various late fusion approaches. Note that, the numbers shown below the `keyword-driven object' column in the table do not indicate the detection/classification performances for the corresponding objects. Whenever an object detectors/classifier finds the corresponding object in a given image, the image is classified as `malicious'. Thus, when the {\it police} detector detects a police officer in a test image which is originally labeled as `benign', this instance is counted as a false positive for that detector.

Keyword-driven object detectors/classifiers do not provide better classification accuracy than the baseline. This is because these semantically relevant objects are only seen in small portions in the dataset. However, it is interesting to notice that the accuracy can be boosted up consistently across different fusion methods when the baseline classifier is combined with these keyword-driven classifiers/detectors.\\

\begin{table}[t]
\setlength{\tabcolsep}{4.5pt}
\begin{center}
\caption{{\bf Overall Performance Comparison.} Early- and late fusion approaches}
\begin{tabular}{l|c|c|c}
\hline
Method & Keyword Info. & AP & Gain 
\\\hline\hline
Event CNN & No & 72.2 & \\
+ Late fusion & Yes & \underline{{\bf 77.1}} & +4.9 \\\hline
Event CNN+ \cite{SEumICIP17} & No & 90.2 & \\
IOD-CNN (Early fusion) \cite{SEumICIP17} & Yes & 93.6 & +3.4 \\
IOD-CNN + Late fusion \cite{SEumICIP17} & Yes & \underline{{\bf 94.2}} & +4.0 \\\hline
\end{tabular}
\label{tab:overallComparison}
\end{center}
\end{table}



\noindent{\bf Keyword Information Helps Consistently.} Table \ref{tab:overallComparison} shows the event classification performance acquired by the selected fusion approaches. The second row in the table indicates the performance gained over the first baseline (`Event CNN') by using the late fusion of separately learned classifiers (see the baseline and the fusion result in Table \ref{tab:ap}). The next three rows show the classification accuracy based on a different baseline classifier (`Event CNN+'). This baseline also does not exploit any keyword information and is reported \cite{SEumICIP17} to have used additional treatments such as an ROI pooling and a different training scheme. IOD-CNN \cite{SEumICIP17} which embeds the keyword-driven object information by early-fusion outperforms its baseline (`Event CNN+') by 3.4 AP. When a late fusion is added on top of that, additional performance increase of 0.6 AP is acquired. See \cite{SEumICIP17} for detailed description.

The results consistently show that exploiting the keyword-driven object detectors/classifiers provides beneficial information in boosting up the event classification performance.

\section{Conclusions}
We addressed a challenging classification problem where certain classes can be expressed by similar visual attributes but should be distinguished from each other semantically. To demonstrate, we have constructed a novel Malicious Crowd Dataset with images representing two classes (benign and malicious) that may look similar but are semantically different. To provide additional semantic information, we collected a set of semantic keywords using a crowd-sourcing platform. We have confirmed the validity of these keywords by analyzing the attention map visualizations for the na\"ive event classifiers (`Event CNN'). We also provide an empirical study which shows the practicality of using semantic keyword information in enhancing the malicious event classification performance.







\bibliographystyle{IEEEbib}
\bibliography{references}

\end{document}